\documentclass[10pt,twocolumn,letterpaper]{article}

\usepackage{cvpr}   
\usepackage{multirow}

\usepackage[dvipsnames]{xcolor}

\definecolor{cvprblue}{rgb}{0.21,0.49,0.74}
\usepackage[pagebackref,breaklinks,colorlinks,citecolor=cvprblue]{hyperref}

\def\paperID{11970} 
\def\confName{CVPR}
\def\confYear{2025}

\title{Efficient Event-Based Object Detection: A Hybrid Neural Network with Spatial and Temporal Attention}

\author{
Soikat Hasan Ahmed\thanks{Conceptual design, algorithm development and experimentation.}, 
Jan Finkbeiner\thanks{Conceptual design, hardware analysis and deployment.}, 
Emre Neftci \\
\textnormal{Forschungszentrum J\"ulich, RWTH Aachen University} \\ 
{\tt\small \{s.ahmed, j.finkbeiner, e.neftci\}@fz-juelich.de}
}

\begin{document}
\maketitle
\begin{abstract}
Event cameras offer high temporal resolution and dynamic range with minimal motion blur, making them promising for robust object detection.  While Spiking Neural Networks (SNNs) on neuromorphic hardware are often considered for energy efficient and low latency event-based data processing, they often fall short of Artificial Neural Networks (ANNs) in accuracy and flexibility. 
Here, we introduce Attention-based Hybrid SNN-ANN backbones for event-based object detection to leverage the strengths of both SNN and ANN architectures. 
A novel Attention-based SNN-ANN bridge module captures sparse spatial and temporal relations from the SNN layer and converts them into dense feature maps for the ANN part of the backbone. Additionally, we present a variant that integrates DWConvLSTMs to the ANN blocks to capture slower dynamics. This multi-timescale network combines fast SNN processing for short timesteps with long-term dense RNN processing, effectively capturing both fast and slow dynamics.
Experimental results demonstrate that our proposed method surpasses SNN-based approaches by significant margins, with results comparable to existing ANN and RNN-based methods. Unlike ANN-only networks, the hybrid setup allows us to implement the SNN blocks on digital neuromorphic hardware to investigate the feasibility of our approach.
Extensive ablation studies and implementation on neuromorphic hardware confirm the effectiveness of our proposed modules and architectural choices.
Our hybrid SNN-ANN architectures pave the way for ANN-like performance at a drastically reduced parameter, latency, and power budget. 
\end{abstract}

\section{Introduction}
\label{sec:intro}

Over the past decade, deep learning has made significant advances in object detection. State-of-the-art approaches predominantly rely on frame-based cameras which capture frames at a fixed rate.  Frame cameras provide dense intensity data but have limitations in dynamic range and frame rates, leading to motion blur. Dynamic Vision Sensors (DVS), or event cameras, offer an alternative by asynchronously capturing pixel-level illumination changes, achieving low latency ($\sim$10$\mu$s), higher temporal resolution, and an extended dynamic range (140 dB vs.\ 60 dB) \cite{gallego2020event}. These characteristics make them well-suited for low-light and fast-motion scenarios. However, due to the sparse, high-temporal-resolution data they generate, effectively processing event data for object detection remains a challenging and emerging research area. Early adopters of event-based object detection ANN models often naively repurpose architectures originally designed for frame-based cameras 
\cite{schaefer2022aegnn,messikommer2020event,iacono2018towards,chen2018pseudo,yik2023neurobench}. ANN models generally achieve good accuracy but tend to be large in terms of parameter count and MAC operations, making them less suitable for deployment on power-efficient edge or neuromorphic devices. Furthermore, the high sparsity and temporal resolution is often discarded in favor of dense representations to leverage GPUs' dense, vector-based representations. 
In contrast, SNNs implemented in neuromorphic hardware are ideally suited to leverage the sparsity of event-based inputs, offering significant reductions in computational cost, power consumption, and latency \cite{davies2018loihi,barchi2019flexible,sawada2016truenorth}. However, SNNs tend to be less accurate at the task level compared to their ANN counterparts.

In this work, we create a hybrid SNN-ANN-based backbone architecture to combine the efficient, event-driven processing of SNNs on neuromorphic hardware with the efficient learning and representation capabilities of ANNs. The SNN extracts low-level features with high temporal resolution from the event-based sensors and converts them into intermediate features, which then change to slower timescales before being processed by the ANN with dense activations. 
Additionally, we feature a variant that adds DWConvLSTMs~\cite{shi2015convolutional, gehrig2023recurrent} to the ANN block. This multi-timescale RNN variant combines the sparse SNN processing of short timesteps with long time horizon processing via the dense RNN with the extracted long timesteps to efficiently capture both fast and slow dynamics of the data. 
For hybrid models, the SNN can be efficiently deployed at the edge in a power efficient manner, as we demonstrate with an implementation on a digital neuromorphic chip. ANN processing can occur either at the edge or, with reduced data rates, in a cloud setting. 
By training the network jointly, the SNN component can leverage backpropagated errors for efficient training via the surrogate gradient approach \cite{Neftci_etal19_surrgrad}.

Information in SNNs is communicated by spike events. In our hybrid models, these must be efficiently converted into dense representations without discarding valuable spatiotemporal features. 
In our model, this is achieved by an attention-based SNN-ANN bridge module, which bridges SNN representations to ANN representations.

The attention module contains two attention modules named Event-Rate Spatial (ERS) attention and Spatial Aware Temporal (SAT) attention. The SAT attention module addresses the challenge of sparse event inputs by enhancing the model’s understanding of irregular structures and  temporal attention to discern temporal relationships within the data. On the other hand, the ERS attention module focuses on highlighting spatial areas by leveraging the activity of events.
Moreover, we implement the SNN blocks in digital neuromorphic hardware to demonstrate the feasibility of our approach.
We report the performance of our model on large-scale event-based object detection benchmarks.  The contributions of our work can be summarized as follows:
\begin{itemize}
    \item A novel hybrid backbone-based event object detection model. To the best of our knowledge, this is the first work to propose a hybrid object detection approach for benchmark object detection task. Evaluation on the Gen1 and Gen4 Automotive Detection datasets \cite{de2020large, perot2020learning} shows that the proposed method outperforms SNN-based methods and achieves comparable results to ANN and RNN-based approaches.
    \item An attention-based SNN-ANN bridge module ($\beta_{asab}$) to convert spatiotemporal spikes into a dense representation, enabling the ANN part of the network to process it effectively while preserving valuable spatial and temporal features through the Event-Rate Spatial (ERS) and Spatial-Aware Temporal (SAT) attention mechanisms.
     \item A multi-timescale RNN variant that includes both the high-temporal resolution SNN block followed by a slower, long time horizon DWConvLSTMs in the ANN block, operating on larger timesteps extracted via the $\beta_{asab}$ module.
    \item Implementation of the SNN blocks on digital neuromorphic hardware to validate its performance and efficiency.

\end{itemize}

\section{Related Work}
Recent studies demonstrated the potential of event cameras in object detection tasks. 
In the earlier stages of adopting event cameras, the focus primarily revolved around adapting existing frame-based feature extractors and a detection head for object detection using event data \cite{iacono2018towards, chen2018pseudo}.
In \cite{iacono2018towards}, researchers integrated event-based data into off-the-shelf frame-based object detection networks. They employed an InceptionNet-based backbone for feature extraction and a single-shot detector (SSD) for detection \cite{szegedy2015going,liu2016ssd}. Similarly, \cite{perot2020learning} utilized a frame-based object detection model called RetinaNet, which incorporates a spatial pooling-based feature extractor \cite{lin2017feature} along with a detection head, applied to event data.

Additionally, methods such as \cite{gehrig2023recurrent, perot2020learning, li2022asynchronous} have incorporated recurrent neural networks (RNNs) as feature extractors for event data. \cite{zubic2024state} uses SSM to improve training time, and \cite{wu2024leod} proposes a training schema with efficient ground truth label utilization.
\cite{messikommer2020event} introduced sparse convolution as a method for event feature extraction. To address the challenges of efficiently extracting spatiotemporal features, \cite{schaefer2022aegnn} investigates the usability of a graph neural network-based approach as a feature extractor. 

Recently, SNN-based methods have become popular for event data processing due to their spike-based working principle, similar to event cameras, which enables efficient processing.
Research conducted by \cite{cordone2022object} and \cite{su2023deep} showcases the effective utilization of SNNs in object detection tasks. Specifically, \cite{cordone2022object} and \cite{su2023deep} delve into assessing the performance of converting widely-used ANN-based backbone architectures such as SqueezeNet~\cite{iandola2016squeezenet}, VGG~\cite{simonyan2014very}, MobileNet~\cite{howard2017mobilenets}, DenseNet~\cite{huang2017densely}, and ResNet~\cite{he2016deep} into SNN architecture for event-based object detection. Nonetheless, optimizing intermediate and high-level features for detection with SNNs results in a significant drop in accuracy.

Recognizing the distinct advantages offered by both SNNs and ANNs, researchers have explored merging these networks into hybrid architectures \cite{kugele2021hybrid}.
For instance, \cite{zhao2022framework} presents a framework that leverages hierarchical information abstraction for meta-continual learning with interpretable multimodal reasoning. Building on this idea, \cite{yang2019dashnet} introduces DashNet, which integrates SNNs with ANN-based feature extraction for high-speed object tracking. Similarly, \cite{liu2022enhancing} improves SNN performance through a hybrid top-down attention mechanism, while \cite{lee2020spike} demonstrates that hybrid models can achieve energy-efficient optical flow estimation with enhanced robustness. Complementing these advances, \cite{kugele2021hybrid, Aydin_2024_CVPRW} develops an architecture that fuses SNN backbones with ANN heads for event-based vision tasks.
By leveraging the complementary strengths of each, these hybrid networks show promise for simpler tasks. However, the bridge between SNNs and ANNs is still overlooked to harness the best of both worlds.

Moreover, the full extent of their capabilities remains largely unexplored, especially in tackling state-of-the-art benchmark vision tasks, such as object detection on popular datasets like Gen1 Automotive Detection dataset \cite{de2020large} and Gen4 Automotive Detection dataset \cite{perot2020learning}.

\section{Hybrid Object Detection Network}
\label{sec:method}

\begin{figure*}[htp]
    \centering
    
    \begin{subfigure}{0.8\textwidth}
        \centering
        \includegraphics[width=\linewidth]{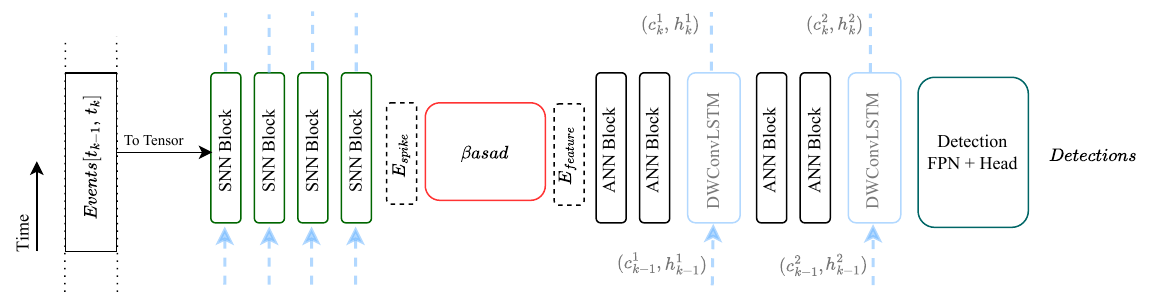}
    \end{subfigure}
    
    \caption{Architecture of the hybrid model with object detection head and SNN-ANN hybrid backbone, including the SNN part, $\beta_{asab}$ bridge module and ANN part. The DWConvLSTM modules and dashed blue arrows are only part of the proposed hybrid + RNN variant.}
    \label{figure:overall}
\end{figure*}
The overall hybrid network as shown in Figure \ref{figure:overall} comprises two key parts: an attention-based hybrid backbone designed to extract spatio-temporal features, and detection heads tasked with identifying objects. In the following section, we will delve into the details of the core components of the network.

\subsection{Event Representation}
An event is represented as $e_n = (x_n, y_n, t_n, p_n)$, where $(x_n, y_n)$ is the pixel location, $t_n$ is the time, and $p_n$ is polarity which indicates the change in light intensity (i.e., positive or negative). 
The event data is pre-processed to convert it into a time-space tensor format.
Following \cite{gehrig2023recurrent}, we start by creating a 4D tensor $\textbf{ $Events[t_{k-1}, t_k]$}  \in \mathbb{R}^{T \times 2 \times H \times W}$, where $T$ represents number of time discretization steps, $2$ denotes polarity features which contain the count of positive and negative events in each discretized time step, and $H$ and $W$ signify the height and width of the event camera, respectively.
Given the event set $\mathcal{E} = \{ e_1, e_2, \dots, e_N \}$, the event tensor $\textbf{ $Events[t_{k-1}, t_k]$} $ is constructed from the discretized time variable $t^\prime_n=\left\lfloor\frac{t_n-t_a}{t_b-t_a} \cdot T\right\rfloor$ as follows:
\begin{equation}
\begin{split}
\textit{Events}[t_{k-1}, t_k](t, p, x, y) &= \sum_{e_n \in \mathcal{E}} \delta\left(p - p_n\right) \delta\left(x - x_n\right) \\
& \quad \delta\left(y - y_n\right) \delta\left(t - t^\prime_n\right).
\end{split}
\end{equation}
While training, event tensors are created. However, during inference, given an input sparsity of $\sim 98\%$ (for Gen 1), this results in significant efficiency gains due to sparse processing in neuromorphic hardware compared to the dense processing in a GPU.
\subsection{Attention-based Hybrid Backbone}
The proposed hybrid backbone architecture, as shown in Figure \ref{figure:overall}, consists of three fundamental components: a low-level spatio-temporal feature extractor $f_{snn}$, an ANN-based high-level spatial feature extractor $f_{ann}$, and a novel Attention-based SNN-ANN Bridge (ASAB) module $\beta_{asab}$.

The first module, denoted as $f_{snn}$, is an event-level feature extractor operating in the spatio-temporal domain and consists of multiple convolutional SNN blocks. Each block follows a structured sequence of operations: standard convolution, batch normalization \cite{ioffe2015batch}, and Parametric Leaky Integration and Fire (PLIF) spiking neuron \cite{Fang_2021_ICCV}. The neural dynamics of PILF with trainable time constant $\tau = \text{sigmoid}(w)^{-1}$ given input $X[t]$ can be expressed as follows:
\begin{equation}
     V[t] = V[t-1] + \frac{1}{\tau}(X[t] - (V[t-1] - V_{reset}) \label{equ:plif} .
\end{equation}
The $f_{snn}$ module receives $\textbf{ $Events[t_{k-1}, t_k]$} $ as its input and generates events
$\textbf{E}_{spike} = f_{snn}(\textbf{ $Events[t_{k-1}, t_k]$} )\in \mathbb{R}^{T \times C \times H' \times W'}$. 
As SNNs operate on a faster timescale and utilize sparse representations and ANNs operate on dense representations, efficiently translating valuable spatio-temporal information into dense representations is essential. To achieve this translation, the $\textbf{E}_{spike}$ is subsequently fed into a proposed $\beta_{asab}$ module, which bridges the SNN and the ANN parts. The events $\textbf{E}_{spike}$ are converted into dense, non-binary features while preserving spatial and temporal information in the form of spatial feature maps. The output of $\beta_{asab}$ is represented by $\textbf{F}_{out}=\beta_{asab}(\textbf{E}_{spike})$, with dimensions $C \times H' \times W'$ which is compatible with traditional 2D convolution-based networks, allowing for smooth processing and integration of information across both spatial and temporal dimensions. The attention module is further described in Section \ref{sec:attention}.

The third component, $f_{ann}$, extracts high-level spatial features using multiple ANN blocks with standard ANN components. Each ANN block consists of standard convolution operations, normalization \cite{ioffe2015batch,ba2016layer}, and ReLU activation functions, enabling the extraction of detailed high-level spatial features from the densely encoded $\textbf{F}_{out}$. 

In addition to the proposed model, we explore a variant that features an added RNN module, incorporating two Depth-Wise separable Convolutional LSTM (DWConvLSTM) units similar to those in \cite{gehrig2023recurrent}, as illustrated in Figure \ref{figure:overall}.

The $f_{snn}$ processes fast dynamics with small timesteps from the event-based camera, while the DWConvLSTM operates on larger timesteps extracted from the $\beta_{asab}$-module to capture slower dynamics.
The resulting outputs from the ANN blocks are then fed to the detection head for the final object detection output. 
%
\begin{figure*}[!ht]
	\centering
        \begin{subfigure}{.59\linewidth}
            \includegraphics[width=\linewidth]{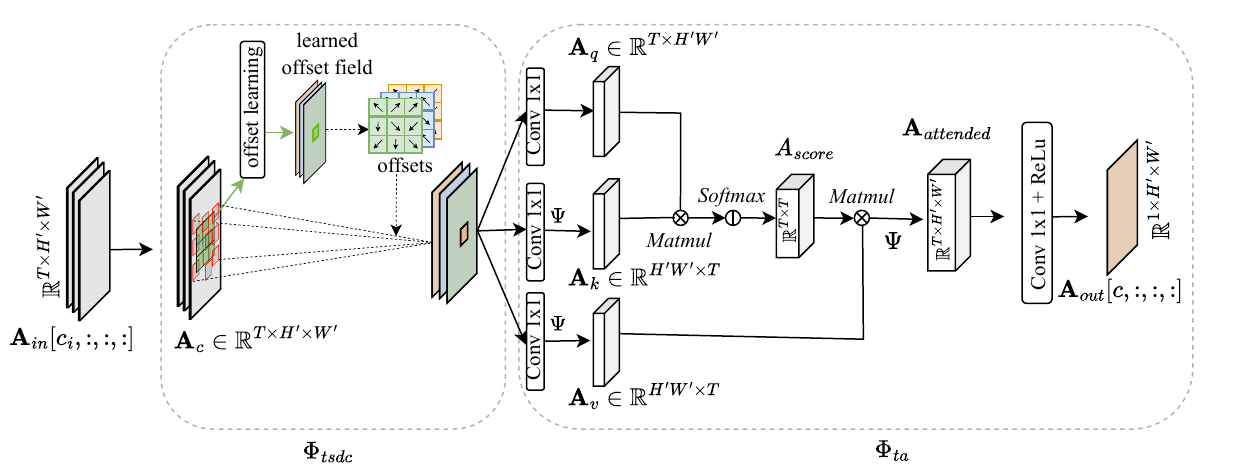}
            \caption{Spatial-aware Temporal Attention.}
                \label{figure:atten_1}
        \end{subfigure}
	\begin{subfigure}{.39\linewidth}
            \centering
            \includegraphics[width=\linewidth]{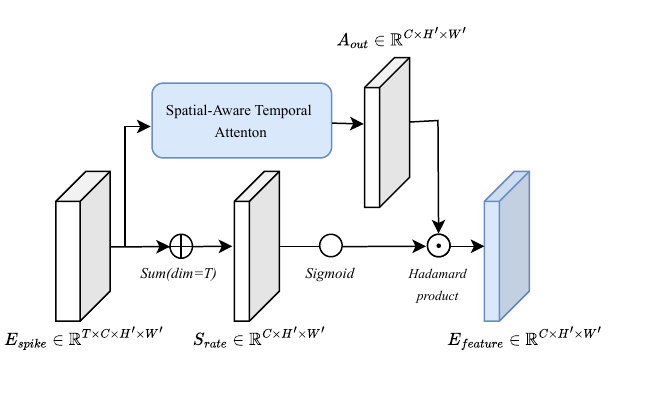}
             \caption{Event-rate Spatial Attention.}
            \label{figure:atten_2}
	\end{subfigure}
\caption{Visualization of the proposed attention module's components. (a) Spatial-aware Temporal Attention: Highlights relevant temporal features in spatial regions to enhance temporal coherence in event-based data. (b) Event-rate Spatial Attention: Emphasizes spatial regions based on event rates, allowing for adaptive focus on areas with significant event activity. Together, these components improve feature extraction in spatiotemporal data processing.}
\label{figure:attenion}
\end{figure*}
\subsection{Attention-based SNN-ANN Bridge Module}
\label{sec:attention}
The bridge module $\beta_{asab}$ comprises two attention modules: i) Spatial-aware Temporal (SAT) attention and ii) Event-Rate Spatial (ERS) attention. The SAT attention module dynamically captures local spatial context within the irregular spatial spike-structure to uncover temporal relations. Meanwhile, the ERS attention submodule focuses on attending to spatial areas utilizing the spatial event activities. Below, we describe these two submodules.
\subsubsection{Spatial-aware Temporal (SAT) Attention}
\label{sec:sat}
The SAT attention contains three crucial operations: i) Channel-wise Temporal Grouping to group relevant features from different time dimensions, ii) Time-wise Separable Deformable Convolution (TSDC) denoted as $\Phi_{tsdc}$ for capturing channel-independent local spatial context from sparse spike features, and iii) Temporal attention module $\Phi_{ta}$, which uses local spatial context features to extract temporal relations to accumulate and translate temporal information into spatial information.

%
\paragraph{Time-wise Separable Deformable Convolutions (TSDC):}
At first, we apply the channel-wise temporal grouping operation to the input data so that each feature channel is processed separately while capturing spatial and temporal relations. This operation transforms the input spike tensor $\textbf{E}_{spike} \in \mathbb{R}^{T \times C \times H' \times W'}$ into $\textbf{A}_{in} \in \mathbb{R}^{C \times T \times H' \times W'}$.  

As shown in Figure \ref{figure:atten_1}, the $\Phi_{tsdc}$ operation operates on individual channel-wise temporal groups $A_{in}[c_i, :, :, :]$, denoted as $A_{c}$, where $i$ is the channel index.  This operation extracts the local spatial context of the sparse, irregularly shaped spike-based representations. We posit that this irregular representation is better extracted using a deformed kernel rather than a standard square grid kernel, as discussed in Section \ref{sec:asab_module}.
We implemented the TSDC as a time-wise separable convolution to capture spatial details independently of time-based changes, as motion occurs over time. Isolating spatial aspects enables a clearer understanding of the structure and layout of features, separate from their movement.

The time-wise separated spatial context is then passed to the $\Phi_{ta}$ module for further processing to determine the temporal relation of different time dimensions.

For the implementation of TSDC, we utilize deformable convolution introduced by \cite{dai2017deformable} which adjusts sampling points on the standard grid by dynamically predicting kernel offsets based on input features. During training, an additional convolutional layer called "offset learning" (refer to  Figure \ref{figure:atten_1}) is trained to predict these offsets. 
Moreover, to independently process each temporal dimension, we set the group of deformable convolution kernels equal to the number of time steps $T$. This is done to encourage the network to focus on the spatial context of the data while maintaining temporal relations intact for further processing.
\paragraph{Temporal Attention:}
To learn relationships between different time steps, we pass the local spatial context $\textbf{A}_{sc}=\Phi_{tsdc}(A_c)$ through a temporal attention module. 
This module leverages the multi-head softmax self-attention mechanism introduced in \cite{vaswani2017attention}. 
In our case, we apply self-attention along the temporal dimension to extract temporal relations.

Firstly, we calculate the keys, queries, and values for temporal self-attention by employing $1 \times 1$ convolutions, followed by a reshape operation, which we denote as $\textbf{A}_k$, $\textbf{A}_q$, and $\textbf{A}_v$, respectively. These operations output tensors of shapes $\mathbb{R}^{H'W' \times T}$, $\mathbb{R}^{T \times H'W'}$, and $\mathbb{R}^{T \times H'W'}$, respectively.
\begin{equation}
    \textbf{A}_{k} = \omega_k(A_{sc}), \quad \textbf{A}_{q} = \Psi(\omega_q(A_{sc})), \quad \textbf{A}_{v} = \Psi(\omega_v(A_{sc}))
\end{equation}
Where $\Psi$ denotes the reshape operation and $\omega$ denotes the $1 \times 1$ convolution operation.
Next, the temporal attention scores denoted as $\textbf{A}_{score} \in \mathbb{R}^{T \times T}$ are computed by performing matrix multiplication between $\textbf{A}_{q}$ and $\textbf{A}_{k}$, followed by applying a $\text{softmax}$ operation:
\begin{equation}
\label{equ:ascore}
    \textbf{A}_{score} = \text{softmax} \left( \textbf{A}_q \textbf{A}_k \right) .
\end{equation}
To obtain the attended temporal features, $A_{v}$ is multiplied with $A_{score}$, followed by a reshape operation to output $A_{attended} \in \mathbb{R}^{T \times H' \times W'}$:
\begin{equation}
\label{equ:attended}
    \textbf{ A}_{attended}= \Psi(\textbf{A}_v \textbf{A}_{score}).
\end{equation}
Finally, a weighted-sum along the temporal dimension using a $1 \times 1$ convolution produces the output $\textbf{A}_{out}[c, :, :] \in \mathbb{R}^{H' \times W'}$. This operation effectively combines the attended temporal features to produce the final output.
\subsubsection{Event-rate Spatial Attention}
\label{sec:ers}
This attention module extracts spatial correlation as spatial weights, utilizing dynamic event activity from intermediate spikes generated by the $f_{snn}$ module.
To identify active regions, an Event-rate Spatial Attention mechanism takes the input $\textbf{E}_{spike}$, and sums the time dimension to calculate the event rates $\textbf{S}_{rate}$, resulting in a shape of $\mathbb{R}^{C\times H' \times W'}$: $\textbf{S}_{rate} = \sum_{t=1}^{T} E_{spike}(t, :, :, :)$.
The $\textbf{S}_{rate}$ is first normalized using a sigmoid function to provide a spatial attention score based on the event activity. This attention score is then utilized as a weight to adjust the output of the SAT module through a Hadamard product, as visualized in Figure \ref{figure:atten_2}:
\begin{equation}
\label{equ:gate}
\textbf{E}_{feature} = \text{sigmoid}(\textbf{S}_{rate}) \odot \textbf{A}_{out}
\end{equation}
The resulting tensor $F_{out}$ is then fed into ANN blocks, which are subsequently utilized to predict the object detection bounding box by a detection head \cite{ge2021yolox}.
\section{Experiments}
\label{sec:experiments}
\subsection{Setup}
\paragraph{Datasets:} 
To conduct the training and evaluation of our network, we utilized two event-based object detection datasets: Gen1 Automotive Detection dataset \cite{de2020large} and Gen4 Automotive Detection dataset \cite{perot2020learning}. The Gen1 and Gen4 datasets comprise 39 and 15 hours of event camera recordings at a resolution of 304 × 240 and 720 × 1280, respectively, with bounding box annotations for car, pedestrian, and two-wheeler (Gen4 only) classes.
\paragraph{Implementation Details:}
The model is implemented in PyTorch \cite{paszke2019pytorch} with the SpikingJelly library \cite{fang2023spikingjelly} and trained end-to-end for 50 epochs on the Gen 1 dataset and 10 epochs on the Gen 4 dataset. The ADAM optimizer \cite{kingma2014adam} is used with a OneCycle learning rate schedule \cite{smith2019super}, which decays linearly from a set maximum learning rate. The kernel size for $\Phi_{tsdc}$ is set to 5.
The training pipeline incorporates data augmentation methods such as random horizontal flips, zoom, and crop, based on \cite{gehrig2023recurrent}. Event representations for the SNN are constructed from 5 ms bins. During training, object detections are generated every 50 ms, using the SNN’s output from the last 10 time bins, while inference allows higher temporal resolution, bounded by the SNN timestep.
The YOLOX framework \cite{ge2021yolox} is used for object detection, incorporating IOU loss, class loss, and regression loss. For the Gen 1 dataset, models are trained with a batch size of 24 and a learning rate of $2 \times 10^{-4}$, requiring approximately 8 hours on four 3090 GPUs. On the Gen 4 dataset, the batch size is 8 with a learning rate of $3.5 \times 10^{-4}$, taking around 1.5 days on four 3090 GPUs. 

When using an RNN variant, we follow previous methods with a sequence length of 21 for fair comparison. This RNN-based network, trained for 400,000 steps with a batch size of 2, requires approximately 6 days to complete training.
\subsection{Benchmark comparisons}
\label{subsection:comparison_results}

\paragraph{Comparison Design}
\label{subsubsection:comparison_design}
To the best of our knowledge, this work presents the first hybrid object detection model implemented in large-scale benchmark datasets, rendering comparisons to other work challenging.
Therefore, we design our comparison in three setups -  (i) comparison with existing ANN-based methods 
(ii) comparison with SNN-based object detection methods, and (iii) comparison with RNN-based models.

\paragraph{Evaluation Procedure:}
Following the evaluation protocol established in prior studies \cite{perot2020learning,gehrig2023recurrent,cordone2022object}, the mean average precision (mAP) \cite{lin2014microsoft} is used as the primary evaluation metric to compare the proposed methods' effectiveness with existing approaches. Since most methods do not offer open-source code, the reported numbers from the corresponding papers were used.
\begin{table*}[!h]
\begin{center}
\caption{Comparative analysis of various ANN-based models for event-based object detection on the Gen1 \cite{de2020large} and Gen4 \cite{perot2020learning} Automotive Detection datasets, where mAP denotes mAP(.5:.05:.95). $A^*$ suggests that this information was not directly available and estimated based on the publication. }
\begin{tabular}{ccccc}
\toprule
       &         &        & Gen 1 & Gen 4 \\
Models & Type    & Params & mAP& mAP \\
\midrule
AEGNN \cite{schaefer2022aegnn} & GNN & 20M  & 0.16 & -  \\ 
SparseConv \cite{messikommer2020event}& ANN & $133M$ & 0.15  & - \\ 
Inception + SSD \cite{iacono2018towards}& ANN &  $>60M^*$  & 0.3  & 0.34 \\
RRC-Events \cite{chen2018pseudo}& ANN & $>100M^*$   & 0.31  & 0.34\\ 
Events-RetinaNet \cite{perot2020learning}& ANN & 33M   & 0.34  & 0.18 \\ 
E2Vid-RetinaNet \cite{perot2020learning}& ANN & 44M   & 0.27  & .25\\ 
RVT-B W/O LSTM \cite{gehrig2023recurrent}& Transformer & $16.2M^*$   & 0.32  & -\\ \midrule

\textbf{Proposed}  & Hybrid & 6.6M  & 0.35  & .27  \\ 
\bottomrule
\end{tabular}
\label{tab:ANN-table}
\end{center}
\end{table*}
\begin{table}[!ht]
\begin{center}
\caption{Comparative analysis of various SNN-based models for event-based object detection on the Gen1 Automotive Detection dataset .}
\begin{tabular}{cccc}
\toprule
Models             & Type            & Params                & mAP \\ 
\midrule
VGG-11+SDD \cite{cordone2022object}         & SNN             & 13M                      & 0.17            \\
MobileNet-64+SSD \cite{cordone2022object}  & SNN             & 24M                      & 0.15            \\ 
DenseNet121-24+SSD \cite{cordone2022object}& SNN             & 8M                       & 0.19            \\ 
FP-DAGNet\cite{zhang2023automotive}         & SNN             & 22M                      & 0.22            \\ 
EMS-RES10 \cite{su2023deep}         & SNN             & 6.20M                   & 0.27            \\
EMS-RES18 \cite{su2023deep}         & SNN             & 9.34M                    & 0.29            \\ 
EMS-RES34 \cite{su2023deep}         & SNN             & 14.4M                   & 0.31     
\\

SpikeFPN \cite{zhang2024automotive}          & SNN             & 22M                    & 0.22    
\\\midrule 
\textbf{Proposed}  & Hybrid & 6.6M & 0.35   \\ 
\bottomrule
\end{tabular}
\label{tab:snn-table}
\end{center}
\end{table}
%
\begin{table}[!ht]
\begin{center}
\caption{Comprehensive evaluation of different RNN-based models for event-based object detection tasks on the Gen1 Automotive Detection dataset. Here 'TF' denotes Transformer.} 
\begin{tabular}{cccc}
\toprule
Models & Type & Params   & mAP \\
\midrule

S4D-ViT-B \cite{zubic2024state}& TF + SSM &16.5M & 0.46 \\ 
S5-ViT-B \cite{zubic2024state}& TF + SSM &18.2M & 0.48 \\ 
S5-ViT-S \cite{zubic2024state}& TF + SSM &9.7M & 0.47 \\ 
RVT-B \cite{gehrig2023recurrent}& TF + RNN & 19M & 0.47 \\ 
RVT-S \cite{gehrig2023recurrent}& TF + RNN & 10M & 0.46 \\ 
RVT-T \cite{gehrig2023recurrent}& TF + RNN &4M & 0.44 \\ 
ASTMNet \cite{li2022asynchronous}& (T)CNN + RNN & ~100M & 0.48 \\ 
RED \cite{perot2020learning}& CNN + RNN & 24M & 0.40 \\ \midrule

\textbf{Proposed+RNN} & Hybrid + RNN & 7.7M  & 0.43   \\ 
\bottomrule
\end{tabular}
\label{tab:RNN-table}
\end{center}
\end{table}
\paragraph{Comparison design with ANN-based methods:}

The efficacy of the proposed method was evaluated against ANN-based models.
The results presented in Table \ref{tab:ANN-table} provide a compelling comparison of various ANN-based networks and performance on the event-based Gen 1 dataset. Notably, the Proposed hybrid model stands out with only 6.6M parameters, significantly smaller than other models such as SparseConv (133M) and RRC-Events (100M). Despite its compact size, our proposed model achieves an accuracy of 0.35, outperforming larger models like SparseConv, which achieves 0.15, and closely matching the performance of Events-RetinaNet (33M, 0.34). In contrast, for the Gen 4 dataset, the analysis also includes  architectures like Events-RetinaNet and E2Vid-RetinaNet. Events-RetinaNet achieves a lower mean Average Precision (mAP) of 0.18, while E2Vid-RetinaNet performs slightly better with an mAP of 0.25. The AEGNN model, which utilizes a graph neural network approach, achieves an mAP of 0.16 with 20 million parameters; however, its performance is overshadowed by the proposed hybrid model, which achieves an mAP of 0.27 while maintaining a compact size of only 6.6 million parameters. We observe that larger models tend to achieve higher accuracy due to their increased parameter counts. However, their significant size makes them less suitable for deployment on hardware with limited resources, such as edge devices or neuromorphic systems.

\paragraph{Comparison Design with SNN-based Methods:}
The proposed hybrid method was compared against several SNN-based methods, specifically, VGG-11+SDD \cite{cordone2022object}, MobileNet-64+SSD \cite{cordone2022object}, DenseNet121-24+SSD \cite{cordone2022object}, FP-DAGNet\cite{zhang2023automotive}, EMS-RES10 \cite{su2023deep}, EMS-RES18 \cite{su2023deep}, EMS-RES34 \cite{su2023deep} and SpikeFPN \cite{zhang2024automotive}.

Table \ref{tab:snn-table} displays the comparison results with various state-of-the-art SNN-based object detection methods.  More specifically,  VGG-11+SSD and MobileNet-64+SSD achieve mAP values of 0.17 and 0.15 with parameter counts of 13M and 24M, respectively. DenseNet121-24+SSD, with a smaller parameter size of 8M, slightly outperforms these models with an mAP of 0.19. FP-DAGNet and SpikeFPN, both at 22M parameters, attain an mAP of 0.22. The EMS-RES series showcases incremental improvements, with EMS-RES10 (6.2M) achieving 0.27, EMS-RES18 (9.34M) at 0.29, and EMS-RES34 (14.4M) achieving the highest mAP among SNNs at 0.31. In contrast, the Proposed Hybrid Model surpasses all these SNN models with an mAP of 0.35 while maintaining an efficient parameter size of only 6.6M. This superiority can be attributed to our method's incorporation of a hybrid feature extraction approach with both spatial and temporal attention modules, which are lacking in other methods.

\paragraph{Comparison design with RNN and SSM-based methods:}
\label{sub:rnn_comp}
Although the performance of the RNN-based models generally outperforms models with spiking components, this comparison aims to investigate how the proposed hybrid model is comparable to the RNN. We provide comparisons with RNN-based methods such as RED \cite{perot2020learning}, ASTMNet \cite{li2022asynchronous}, RVT-B \cite{gehrig2023recurrent}, RVT-S \cite{gehrig2023recurrent}, RVT-T \cite{gehrig2023recurrent}, S4D-ViT-B \cite{zubic2024state}, S5-ViT-B \cite{zubic2024state}, and S5-ViT-S \cite{zubic2024state}. 
Table \ref{tab:RNN-table} presents a comparison of results obtained with RNN-based models. It can be seen that two of the works, RED~\cite{perot2020learning} and ASTMNet~\cite{li2022asynchronous}, have substantially larger parameter counts and are therefore expected to perform better. RVT~\cite{gehrig2023recurrent} demonstrates good accuracy at a parameter count comparable to the proposed hybrid network.
Our fully recurrent backbone allows high-frequency detection without re-computing recurrent layers, unlike RVT’s non-causal attention which requires re-computation for every prediction. CNNs are more efficient than MLPs at small batch sizes due to higher arithmetic intensity; in our method, less than $5\%$ of MACs are from attention and MLPs versus $67\%$ in RVT which is harder to deploy on energy-efficient edge and neuromorphic hardware.
%
\subsection{Ablation Study}
\begin{figure}[!h]
      
	\centering
	
	\begin{subfigure}{0.32\linewidth}
		\includegraphics[width=\linewidth]{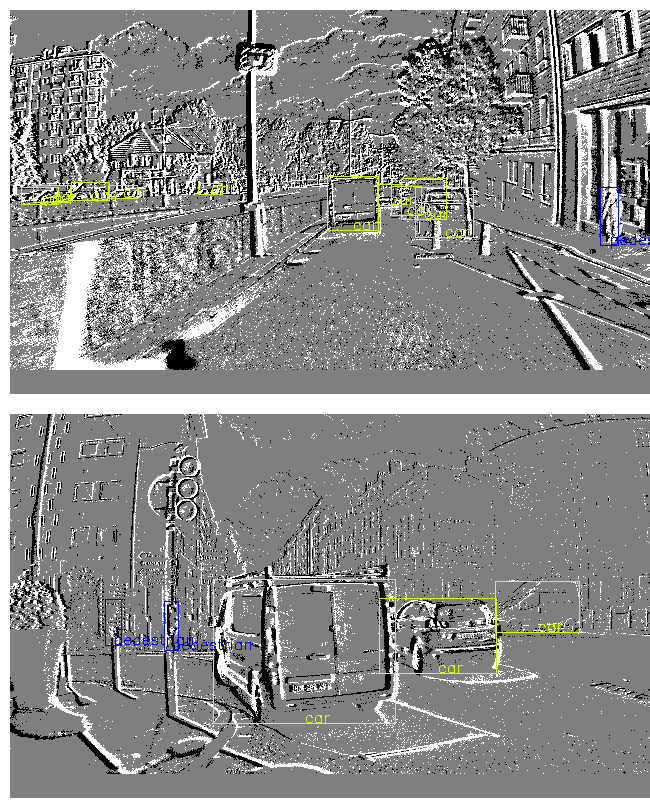}
		\caption{{W/O ASAB module}}
		\label{figure:gen4subfigA}
	\end{subfigure}
	\begin{subfigure}{0.32\linewidth}
		\includegraphics[width=\linewidth]{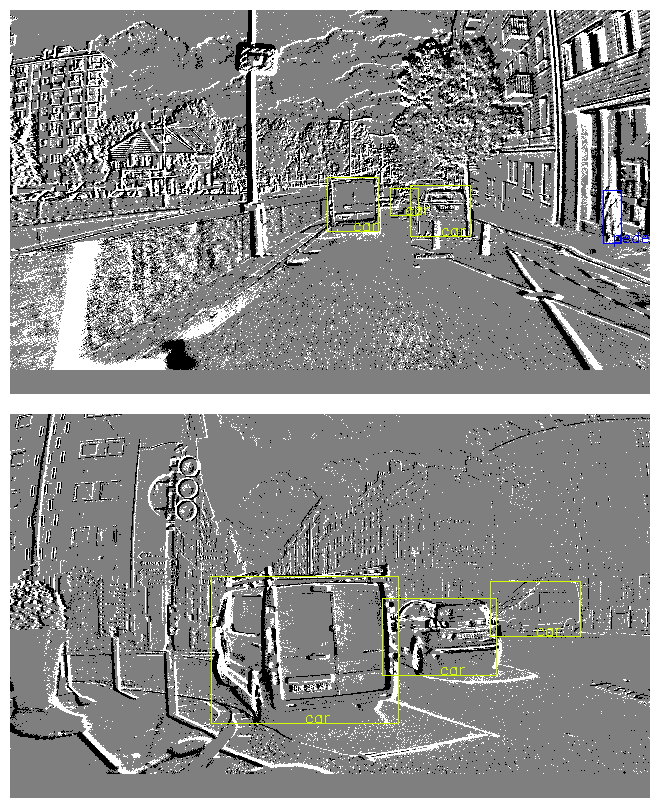}
		\caption{{W/ ASAB module}}
		\label{figure:gen4subfigB}
	\end{subfigure}
	\begin{subfigure}{0.32\linewidth}
	        \includegraphics[width=\linewidth]{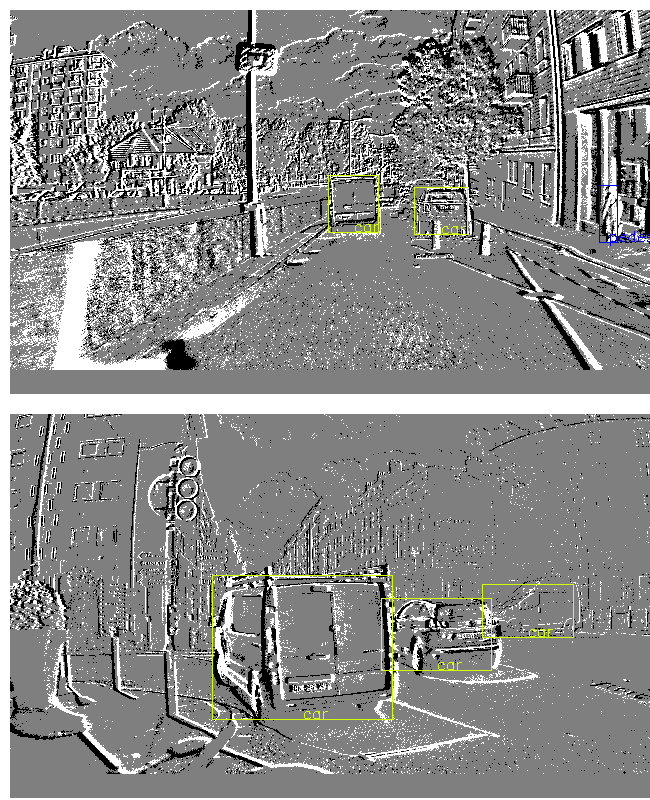}
	        \caption{{Ground Truth}}
	        \label{figure:geen4subfigC}
         \end{subfigure}
\caption{Visual comparison of object detection outputs between the baseline hybrid event object detection method (left) and the proposed method (right) for the Gen 4 dataset. From left to right: (a) object detection output without the ASAB module, (b) object detection output with the ASAB module, and (c) ground truth object boundaries. More samples are in the Supplementary document.}

	\label{figure:vis}
\end{figure}

\begin{table}[!h]
\begin{center}
\caption{Ablation study for ASAB module.}
\begin{tabular}{cccc}
\toprule
Models                   & mAP(.5)       & mAP \\
\noalign{\smallskip}
\midrule
\noalign{\smallskip}
Variant 1(w/o - $\Phi{ta}$)    & 0.57          & 0.33            \\
Variant 2 (w/o deform)& 0.59         & 0.34            \\
Variant 3 (w/o - ESA)          & 0.59          & 0.34            \\
Variant 4 (w/o - ASAB)         & 0.53          & 0.30          \\ 
\textbf{Variant 5} (Proposed)         & 0.61 &\textbf{ 0.35 }  \\ 
\bottomrule
\end{tabular}
\label{tab:ablation_asab}
\end{center}
\end{table}
\paragraph{ASAB module:}
\label{sec:asab_module}
The ablation study highlights the importance of each ASAB module component in enhancing model accuracy on the Gen1 Automotive Detection dataset. In Table \ref{tab:ablation_asab}, Variant 1 (excluding $\Psi_{ta}$) achieves an mAP of 0.33, revealing reduced temporal capture. In variant 2, we replaced the deformable convolution with a standard convolution, which shows irregular sampling helps with sparse data.  In variant  3, removing the ERS module shows  some accuracy drop, indicating limited spatial flexibility and attention. Variant 4, replacing ASAB with a simple accumulation operation, results in the lowest mAP of 0.30. The complete model (Variant 5) reaches the highest mAP of 0.35, emphasizing the value of each component. Figure \ref{figure:vis} illustrates how the bridge module enhances detection by reducing false predictions.
Additionally, we performed an ablation study on various DWConvLSTM configurations by toggling layers. The proposed setting achieved a mAP of 0.43. Please refer to the supplementary (Table \ref{tab:ablation_rnn}) for more details .

\section{Hardware Implementation, Energy and Computational Efficiency Analysis}

\subsection{Hardware Implementation}
\begin{table}[!ht]
\begin{center}
\caption{Power and time measurements of the SNN block on Loihi 2 for several input sizes and number of weight bits. The power is measured in Watts and the execution time per step is in milliseconds. The mean and standard deviation of the measurements averaged over 12 inputs for a total of 100k steps are reported.}
\small
\begin{tabular}{cccccc}
\toprule
\begin{tabular}[c]{@{\hskip .1in}c@{\hskip .1in}}Weight\\ quant.\end{tabular} & \begin{tabular}[c]{@{\hskip .1in}c@{\hskip .1in}} \# chips\\\end{tabular} & \begin{tabular}[c]{@{\hskip .1in}c@{\hskip .1in}}Power [W]\\\end{tabular} & \begin{tabular}[c]{@{\hskip .1in}c@{\hskip .1in}}Time/Step\end{tabular} \\\midrule
int8 & 6 & 1.73 $\pm$ 0.10 & 2.06 \\ 
int6 & 6 & 1.71 $\pm$ 0.11 & 2.06 \\ 
int4 & 6 & 1.95 $\pm$ 0.33 & 1.16 \\ 
\bottomrule
\end{tabular}
\label{tab:hardware}

\end{center}
\end{table}
In order to demonstrate the suitability of the chosen hybrid SNN-ANN approach for energy efficient inference on the edge, we implemented the SNN backbone in hardware. In the proposed architecture the SNN block transforms sensor data into intermediate representations and therefore underlies the strictest latency requirements. Due to the clear separation between SNN and ANN parts in the model's architecture, the SNN blocks can be implemented in specialized hardware. As hardware, we chose Intel's Loihi 2~\cite{davies2018loihi}, a digital, programmable chip based on event-based communication and computation. 
Only minor adjustments are necessary for execution on Loihi 2: 
 The kernel-weights of the convolutional layers are quantized to \texttt{int8} via a per-output-channel quantization scheme showing no resulting loss in accuracy (mAP: 0.348 (\texttt{float16}) vs 0.343 (\texttt{int8})). The batchnorm (BN) operations and quantization scaling are fused into the LIF-neuron dynamics by scaling and shifting the inputs according to the follwing equations:
\small
\begin{align}
    \mathrm{scale} &= \frac{q_\mathrm{scale} \; \mathrm{weight}_\mathrm{BN}}{ \tau \; \sqrt{\mathrm{Var}_\mathrm{BN} + \varepsilon_\mathrm{BN}} } \label{eq:scale_bn_lif} \\
    \mathrm{shift} &= \left( \mathrm{bias}_\mathrm{conv} - \mathrm{mean}_\mathrm{BN} \right) \frac{\mathrm{weight}_\mathrm{BN}}{ \tau \; \sqrt{\mathrm{Var}_\mathrm{BN} + \varepsilon_\mathrm{BN}} } + \frac{\mathrm{bias}_\mathrm{BN}}{\tau} \label{eq:shift_bn_lif}
\end{align}
\normalsize
where $q_\mathrm{scale}$ is the scaling factor introduced by the quantization and $\tau$ is the PLIF neurons time constant.
Given this approach, spike times are almost exactly reproduced on Loihi 2 compared to the PyTorch \texttt{int8} implementation. For benchmarking purposes, the inputs to the network are simulated with an additional neuron population, due to the current IO limitations of the chip. With this approach the spiking statistics in the input and SNN layers are reproduced. 

Table \ref{tab:hardware} reports power and time measurement results of the 4-layer SNN block running on Loihi 2 for inputs of size (2, 256, 160). 
The network runs at $\left( 1.7 \pm 0.1 \right)\,\text{W}$ and $\left( 1.9 \pm 0.8 \right)\,\text{ms}$ per step, which is faster than real-time in the currently chosen architecture ($5\,\text{ms}$ per step).
These results compare favorably to commercially available chips for edge computing like the NVIDIA Jetson Orin Nano ($7\,\text{W} - 15\,\text{W}$)~\cite{JetsonNanoDatasheet} and demonstrate the suitability of an SNN backbone for event-based data processing.

\renewcommand{\thefootnote}{}

\footnotetext{
This research was funded by the German Federal Ministry of Education and Research (BMBF) under the projects "GreenEdge-FuE" (16ME0521) and "Cluster4Future" (03ZU1106CB). Access to JUWELS~\cite{JUWELS} has been granted by GCS (www.gauss-centre.eu) under project neuroml. We thank Intel for access to Loihi 2.}
\subsection{Computational Analysis}

\begin{table}[t]
\centering
\caption{Comparison of different baselines complexities.}
\begin{tabular}{cccc}
\toprule
\text{Models} & \text{mAP(.5)} & \text{MACs} & \text{ACs} \\
\noalign{\smallskip}
\midrule
\noalign{\smallskip}
 $Baseline_{ann}$                             & 0.61  &  $15.34e9$ & $0.0$ \\
$Baseline_{w/o \,\beta_{asab}}$                           & 0.53  &   $1.18e9$ & $0.97e9$ \\
$\mathbf{Proposed_{w/ \beta_{asab}}}$                           & $0.61$  &   $\mathbf{1.63e9}$ & $\mathbf{0.97e9}$ \\ 
$Proposed_{snn+} $                            & 0.58  &   $0.87e9$ & $ 1.59e9$ \\
\bottomrule
\end{tabular}
\label{tab:baseline}
\end{table}
We trained a variant of our model where the modified PLIF neuron acts like a ReLU with proposed attention module to investigate a comparison between a similar artificial neural network (ANN) and the hybrid network. This variant, $Baseline_{ann}$,  with a mean $15.34 \times 10^9$ multiply-accumulate operations (MACs), makes it resource-intensive and unsuitable for energy-constrained hardware. In contrast, our Proposed hybrid model computes only $1.63 \times 10^9$ MACs, making it more practical for edge devices. The $\beta_{asab}$ module does incur additional operations but leads to significant accuracy improvement (compare to $Baseline_{w/o \,\beta_{asab}}$). Additionally, the spiking neural network (SNN) variant, $Proposed_{snn+}$ (Increasing one SNN layer and reducing one ANN layer), reduces computational demands further to $0.87 \times 10^9$ MACs and is highly efficient on neuromorphic hardware, running significantly faster and with less energy on Intel’s Loihi 2 compared to the dense-activation ANNs. 

We analyze power consumption across methods following \cite{Aydin_2024_CVPRW}. Among SNNs, DenseNet121+SSD uses $0.9$ mJ ($0.0$ MACs, $2.3 \times 10^9$ ACs), while VGG-11+SSD requires $4.2$ mJ ($11.1 \times 10^9$ ACs). In ANNs, Inception+SSD is the most demanding at $19.3$ mJ ($11.4 \times 10^9$ MACs). Events-RetinaNet consumes $5.4$ mJ ($3.2 \times 10^9$ MACs), and RVT-B W/O LSTM requires $3.9$ mJ ($2.3 \times 10^9$ MACs). Our method achieves $1.6 \times 10^9$ MACs, $1.0 \times 10^9$ ACs, and $3.1$ mJ, significantly reducing energy and computational costs compared to most ANNs.

%
\section{Conclusion}
\label{sec:conclustion}
In this work, we introduced a hybrid attention-based SNN-ANN backbone for event-based visual object detection. A novel attention-based SNN-ANN bridge module is proposed to capture sparse spatial and temporal relations from the SNN layer and convert them into dense feature maps for the ANN part of the backbone. 
Additionally, we demonstrate the effectiveness of combining RNNs on multiple timescales: hardware-efficient SNNs for fast dynamics on short timescales with ConvLSTMs for longer timescales that operate on the extracted features of the bridge-module.
Experimental results demonstrate that our proposed method surpasses baseline hybrid and SNN-based approaches by significant margins, with results comparable to existing ANN-based methods. The efficacy of our proposed modules and architectural choices is confirmed through extensive ablation studies. Additionally, we demonstrate the effectiveness of our architectural choice with separate SNN and ANN blocks by implementing the SNN blocks on digital neuromorphic hardware, Intel's Loihi 2. The neuromorphic hardware implementation achieves sub-real-time processing and improved power consumption compared to commercially available edge computing hardware. The achieved accuracy and hardware implementation results pave the way toward a hybrid SNN-ANN architecture that achieves ANN-like performance at a drastically reduced parameter and power budget.

{
    \small
    \bibliographystyle{ieeenat_fullname}
    \bibliography{main, biblio_unique_alt}
}

\clearpage
\setcounter{page}{1}
\maketitlesupplementary

\begin{table}[b]
\begin{center}
\caption{This table shows the effects of accuracy due to the quantization of the weights for the SNN blocks, aimed at making them compatible with neuromorphic hardware.}
\begin{tabular}{cccc}
\midrule\noalign{\smallskip}
Models                   & mAP(.5)       & mAP(.5:.05:.95) \\
\toprule
Variant 1 (float16) & 0.613          & 0.348   \\ 
Variant 2 (int8)    & 0.612          & 0.349   \\
Variant 3 (int6)    & 0.612          & 0.348   \\
Variant 4 (int4)    & 0.610          & 0.347   \\
Variant 5 (int2)    & 0.432          & 0.224   \\
\bottomrule
\end{tabular}
\label{tab:int_quant}
\end{center}
\end{table}

\begin{table}[t]
\begin{center}
\caption{Ablation study for DWConvLSTM module in Gen 1 dataset.}
\begin{tabular}{cccccc}
\toprule
Models              & L5 & L6 & L7 & L8 & mAP \\ 
\midrule

 Variant 1  & \checkmark & \checkmark & \checkmark & \checkmark & 0.42 \\ 
 Variant 2  & $\times$ & \checkmark & \checkmark & \checkmark &  0.42 \\ 
 \textbf{Proposed+RNN} & $\times$ & \checkmark & $\times$ & \checkmark &  \textbf{0.43} \\ 
\bottomrule
\end{tabular}
\label{tab:ablation_rnn}
\end{center}
\end{table}

\subsection*{Network Architecture}

\begin{figure*}[!h]
    \centering
    \includegraphics[width=10cm]{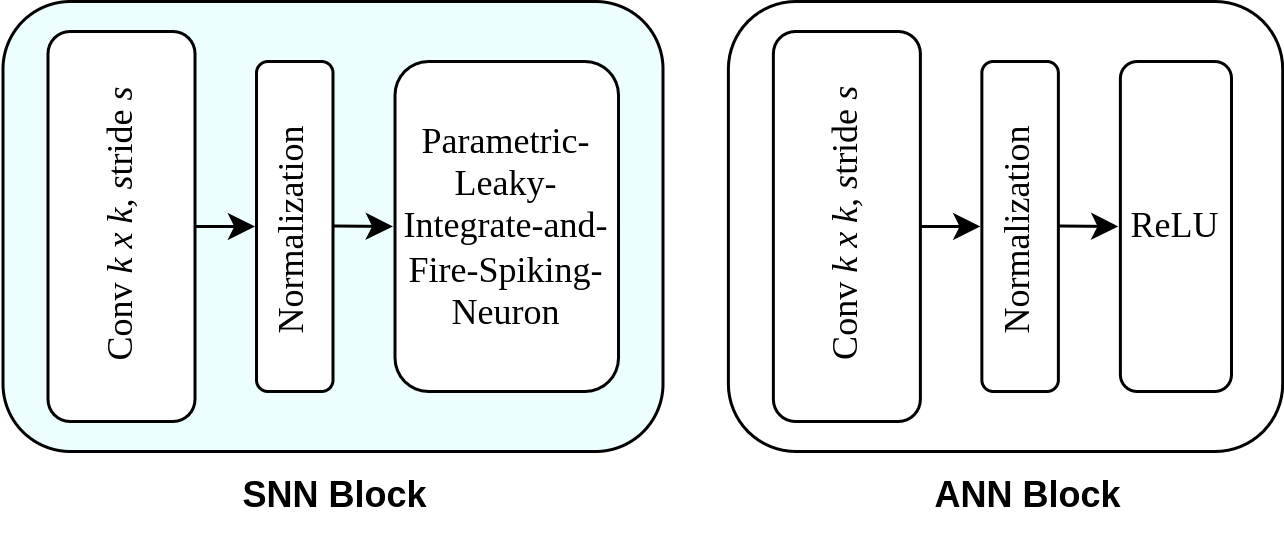}
    \caption{Basic SNN and ANN blocks.}
    \label{fig:blocks}
\end{figure*}

\begin{figure*}[!h]
      
	\centering
	\begin{subfigure}{0.18\linewidth}
		\includegraphics[width=\linewidth]{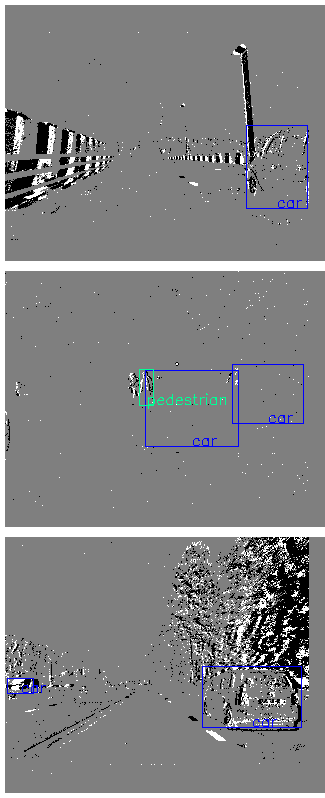}
		\caption{W/O ASAB module}
		\label{fig:gen1subfigA}
	\end{subfigure}
	\begin{subfigure}{0.18\linewidth}
		\includegraphics[width=\linewidth]{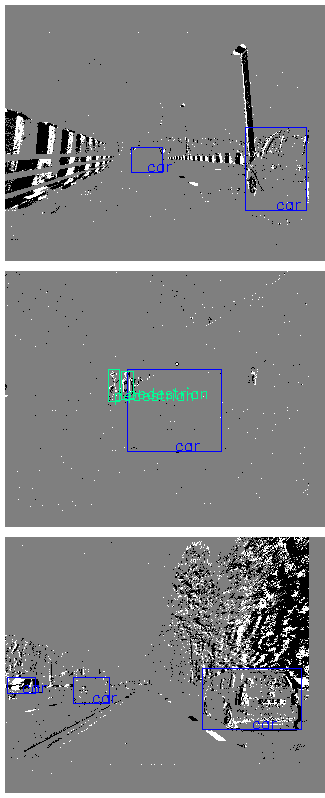}
		\caption{W/ ASAB module}
		\label{fig:gen1subfigB}
	\end{subfigure}
	\begin{subfigure}{0.18\linewidth}
	        \includegraphics[width=\linewidth]{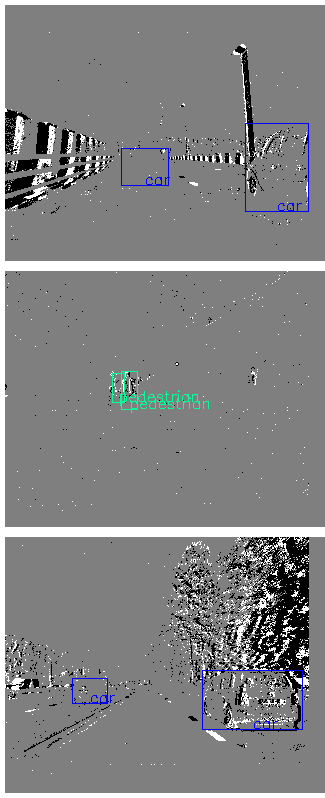}
	        \caption{GT}
	        \label{fig:gen1subfigC}
        \end{subfigure}
        \hfill
	\caption{Visual comparison with the baseline hybrid event object detection method for the Gen 1. From left to right, (a) object detection output of the without proposed bridge module, (b) object detection output of the proposed method (with proposed bridge module), and (c) ground-truth (GT) object boundaries. These visualizations demonstrate that our proposed method significantly improves the detection of smaller objects and mitigates false predictions.}
	\label{fig:vis}
\end{figure*}

Table \ref{tab:backbone_arc_rnn} displays the network architecture of the proposed hybrid backbone. Additionally, Figure \ref{fig:blocks} illustrates the basic SNN and ANN blocks. For better understanding, we have also included an additional figure (Figure \ref{fig:toy}) to clarify the channel-wise temporal grouping operation from Spatial-aware Temporal attention, where similar features from different time dimensions are grouped.
In Figure \ref{fig:vis}, we show additional visual detection outputs from Gen 1 dataset.

\begin{table*}[!ht]
\begin{center}
\caption{Hybrid + RNN architecture with DWConvLSTM.}
\begin{tabular}{cc@{\hskip 0.1in}c@{\hskip 0.1in}c}
\toprule
Layer             & Kernel            & Output Dimensions         & Layer Type   \\ 
\midrule
Input         & - & $ T\times 2\times H\times W$     &    \\ \noalign{\smallskip}
1         & $64c3p1s2$  & $T\times 64\times \frac{1}{2}H\times\frac{1}{2}W$      &    \\ \noalign{\smallskip}
2         & $128c3p1s2$  & $T\times 128\times \frac{1}{4}H\times\frac{1}{4}W$      &  SNN Layers  \\ \noalign{\smallskip}
3         & $256c3p1s2$  & $T\times 256\times \frac{1}{8}H\times\frac{1}{8}W$      &    \\ \noalign{\smallskip}
4         & $256c3p1s1$  & $T\times 256\times \frac{1}{8}H\times\frac{1}{8}W$      &    \\ \noalign{\smallskip} \hline
5         & -  & $ 256\times \frac{1}{8}H\times\frac{1}{8}W$      &   $\beta{asab}$ \\ \noalign{\smallskip} \hline
6         & $256c3p1s1$  & $ 256\times \frac{1}{8}H\times\frac{1}{8}W$      &    \\ \noalign{\smallskip} 
7         & $256c3p1s2$  & $ 256\times \frac{1}{16}H\times\frac{1}{16}W$      &  ANN Layers  \\ \noalign{\smallskip} 
 \midrule
8         & -  & $ 256\times \frac{1}{16}H\times\frac{1}{168}W$      &   DWConvLSTM \\ \noalign{\smallskip} \midrule
9         & $256c3p1s1$  & $ 256\times \frac{1}{16}H\times\frac{1}{16}W$      &    \\ \noalign{\smallskip} 
10         & $256c3p1s2$  & $ 256\times \frac{1}{32}H\times\frac{1}{32}W$      &    \\ \noalign{\smallskip} \midrule
11         & -  & $ 256\times \frac{1}{8}H\times\frac{1}{8}W$      &   DWConvLSTM \\ \noalign{\smallskip} \midrule
          &   & Detection Head  YoloX\cite{ge2021yolox}       &    \\ 
\bottomrule
\end{tabular}
\label{tab:backbone_arc_rnn}
\end{center}
\end{table*}

\subsection*{Hardware Implementation}

This section provides more performance details on the hardware implementation of the spiking layers of the backbone on a digital neuromorphic chip, Intel's Loihi 2 \cite{davies2018loihi}. In Table \ref{tab:power}, we present the Power and time measurements of the SNN block on Loihi 2 for various input sizes, each tested with different weight quantization settings. In Table \ref{tab:int_quant}, effects of accuracy due to the quantization of the weights for the SNN blocks, aimed at making them compatible with neuromorphic hardware.

\begin{table*}[ht]
  \begin{center}
    \caption{Power and time measurements of the SNN block on Loihi 2 for several input sizes and number of weight bits. The power is measured in Watts and the execution time per step in milliseconds. The mean and standard deviation of the measurements averaged over 12 inputs for a total of 100k steps are reported.}
    \label{tab:power}
    \bgroup
    \def\arraystretch{1.2}    
    \begin{tabular}{@{\hskip .01in}ccccc@{\hskip .01in}}
      \toprule
    \begin{tabular}[c]{@{\hskip .1in}c@{\hskip .1in}}Input size\\ (C,W,H)\end{tabular} & \begin{tabular}[c]{@{\hskip .1in}c@{\hskip .1in}}Weight\\ qunatization\end{tabular} & \begin{tabular}[c]{@{\hskip .1in}c@{\hskip .1in}}Number\\ of chips\\\end{tabular} & \begin{tabular}[c]{@{\hskip .1in}c@{\hskip .1in}}Total Power [W]\\\end{tabular} & \begin{tabular}[c]{@{\hskip .1in}c@{\hskip .1in}}Execution Time\\ Per Step [ms]\end{tabular} \\\midrule
  \multirow{3}{*}{(2, 256, 160)} & int8 & 6 & 1.73 $\pm$ 0.10 & 2.06 $\pm$ 0.74 \\ 
 & int6 & 6 & 1.71 $\pm$ 0.11 & 2.06 $\pm$ 0.74 \\ 
 & int4 & 6 & 1.95 $\pm$ 0.33 & 1.16 $\pm$ 0.49 \\ 
\midrule
  \multirow{3}{*}{(2, 128, 160)} & int8 & 4 & 1.51 $\pm$ 0.56 & 0.80 $\pm$ 0.32 \\ 
 & int6 & 4 & 1.81 $\pm$ 0.48 & 0.45 $\pm$ 0.18 \\ 
 & int4 & 3 & 1.39 $\pm$ 0.44 & 0.49 $\pm$ 0.18 \\ 
\bottomrule
    \end{tabular}
    \egroup
  \end{center}
\end{table*}

\subsection*{Channel-wise Temporal Grouping:}
\begin{figure*}[!ht]
	\centering
        \begin{subfigure}{.89\linewidth}
            \includegraphics[width=\linewidth]{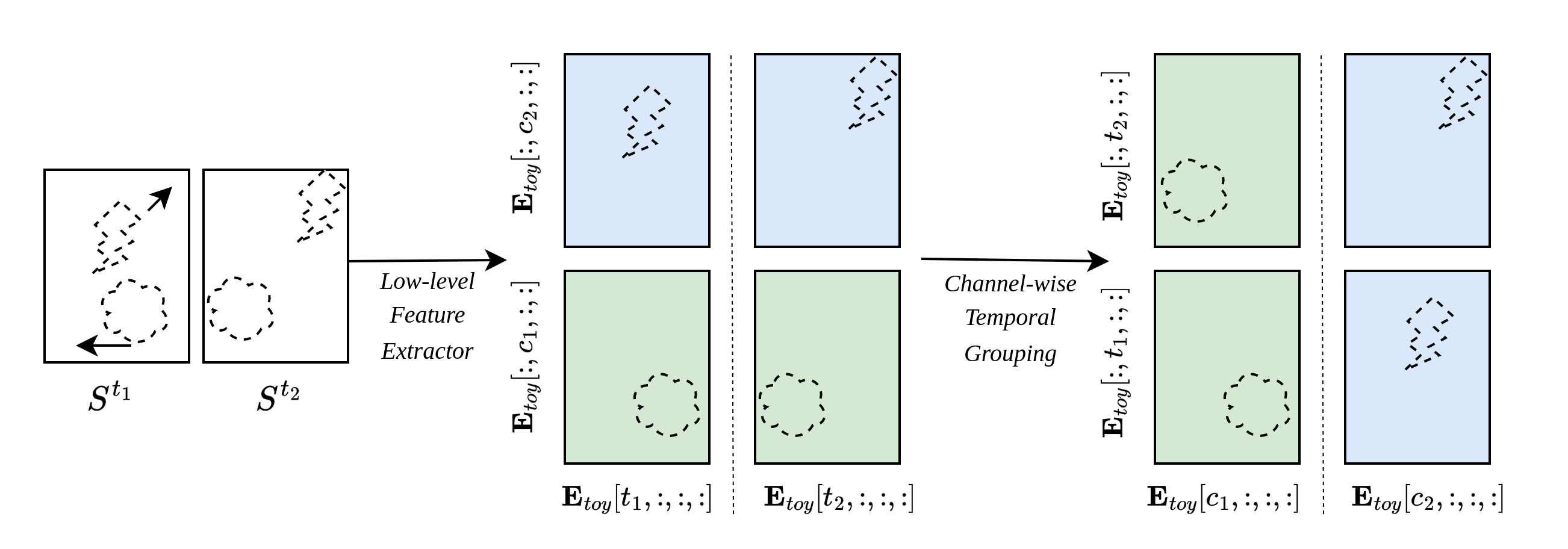}
            \caption{}
            \label{fig:toyctg}
        \end{subfigure}
	\begin{subfigure}{.096\linewidth}
            \centering
            \includegraphics[width=\linewidth]{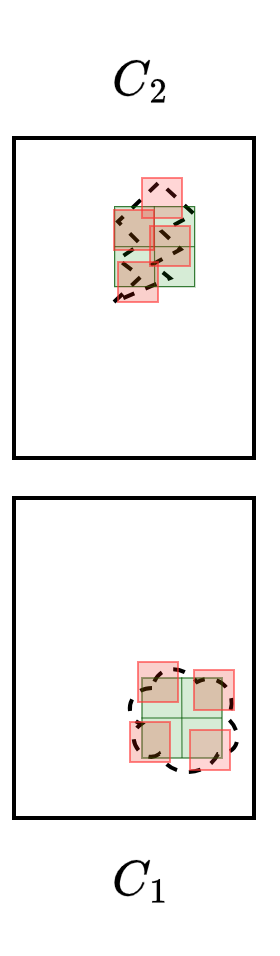}
             \caption{}
            \label{fig:toydeform}
	\end{subfigure}
\caption{Toy examples: Figure \ref{fig:toyctg} illustrates the channel-wise temporal grouping operation. In Figure \ref{fig:toydeform}, deformed kernels highlighted in red are compared with a regular grid marked in green. For visualization, a $2 \times 2$ kernel is shown as an example.}
\label{fig:toy}
\end{figure*}

Consider a scenario where two objects are moving in different directions within a scene. 
Since event cameras detect changes in light intensity, most events captured by the event camera in this scenario will be triggered by the edges of these objects. Additionally, due to their movements over time, there will be spatial shifting, as illustrated in Figure \ref{fig:toyctg}, denoted as $S^{t1}$ and $S^{t2}$.
To extract low-level features from these events, a feature extractor similar to $f_{snn}$ processes the spatio-temporal spikes $\textbf{E}_{toy}$, learning various features such as edges and structures across multiple channels. For illustration, consider two moving features: one round and another lightning-shaped features, as shown in Figure \ref{fig:toy}. We would like to group together events that are produced by one object. This can be accomplished by transposing the $C$ and $T$ dimensions. We call this procedure channel-wise temporal grouping. Note that the input and the output of the feature extractor from the input are simplified in the figure for easier understanding.

\subsection*{Effect of number of SNN blocks}

In Table \ref{tab:ablation_nlayers}, we shows additional experiments to analyze the effect of a number of SNN blocks. Three network variants were examined to assess the impact of different SNN and ANN layer numbers in the proposed architecture. The feature extractor comprised eight layers. Variant 1 decreased SNN layers and increased ANN layers in the $3-4$ setup, resulting in a slight performance boost with mAP(0.75) rising from 0.34 to 0.35. Variant 3, increasing SNN layers in the $5-3$ setup, led to reduced accuracy across all metrics due to fewer ANN layers to extract high-level features. Variant 2, utilizing the $4-4$ setting, balanced between the two, achieving comparable accuracy to Variant 1 with reduced computational overhead from additional ANN blocks. Hence, this configuration was adopted for all subsequent experiments.

\begin{table*}[!h]
\begin{center}
\caption{Ablation study for different settings of the number of SNN and ANN blocks. SNN - ANN represents the number of SNN and ANN blocks in the network.}
\begin{tabular}{cccccc}
\toprule
Models         &  SNN - ANN          & mAP(.5)    & mAP(.75)   & mAP(.5:.05:.95) \\
\midrule
Variant 1   & 3 - 5  & 0.61    &0.35      & 0.35            \\
\textbf{Variant 2}(proposed) & 4 - 4  & 0.61  &0.34 & 0.35   \\ 
Variant 3 & 5 - 3 & 0.58        & 0.33   & 0.33            \\

\bottomrule
\end{tabular}
\label{tab:ablation_nlayers}
\end{center}

\end{table*}

\subsection*{Effect of DWConvLSTMs:}

The ablation study examines configurations of the DWConvLSTM \cite{shi2015convolutional,gehrig2023recurrent} module in the backbone hybrid architecture. In Table  \ref{tab:ablation_rnn}, Variant 1, with all layers (L5-L8), sets a baseline mAP of 0.42. Variant 2, omitting L5 but keeping L6-L8, also achieves an mAP of 0.42, showing L5’s minimal impact. The Proposed+RNN variant, excluding L5 and L7 while adding an RNN with L6 and L8, reaches the highest mAP of 0.43 in the Gen 1 dataset. The RNN variant shows a similar tendency on the  Gen 4 dataset (from 0.27 to 0.34 mAP).

\end{document}